\documentclass[10pt,twocolumn,letterpaper]{article}

\usepackage[pagenumbers]{cvpr}
\usepackage{graphicx}
\usepackage{amsmath}
\usepackage{amssymb}
\usepackage{booktabs}
\usepackage{colortbl}
\usepackage{multirow}
\usepackage{algorithm}
\usepackage{dirtytalk}
\usepackage{cite}
\usepackage{amsfonts}
\usepackage{textcomp}
\usepackage{lipsum}
\usepackage{caption}
\usepackage{soul}
\usepackage{mathtools}
\usepackage{algpseudocode} 
\usepackage{siunitx}
\usepackage{flushend}

\usepackage[pagebackref,breaklinks,colorlinks]{hyperref}

\usepackage[capitalize]{cleveref}
\crefname{section}{Sec.}{Secs.}
\Crefname{section}{Section}{Sections}
\Crefname{table}{Table}{Tables}
\crefname{table}{Tab.}{Tabs.}

\def\confYear{2022}

\usepackage{overpic}
\usepackage{enumitem} 
\usepackage{overpic} 
\usepackage{color}

\definecolor{turquoise}{cmyk}{0.65,0,0.1,0.3}
\definecolor{purple}{rgb}{0.65,0,0.65}
\definecolor{dark_green}{rgb}{0, 0.5, 0}
\definecolor{orange}{rgb}{0.8, 0.6, 0.2}
\definecolor{red}{rgb}{0.8, 0.2, 0.2}
\definecolor{darkred}{rgb}{0.6, 0.1, 0.05}
\definecolor{blueish}{rgb}{0.0, 0.3, .6}
\definecolor{light_gray}{rgb}{0.7, 0.7, .7}
\definecolor{pink}{rgb}{1, 0, 1}
\definecolor{greyblue}{rgb}{0.25, 0.25, 1}

\usepackage{blindtext}

\renewcommand{\paragraph}[1]{\vspace{1em}\noindent\textbf{#1}.}
\begin{document}
\title{Learning Binary and Sparse Permutation-Invariant Representations for Fast and Memory Efficient Whole Slide Image Search}

\author{Sobhan Hemati$\dagger$, Shivam Kalra$\dagger$, Morteza Babaie$\dagger$, H.R. Tizhoosh$\dagger,\ddagger$
\vspace{0.1in}
\\
$\dagger$ Kimia Lab, University of Waterloo, Waterloo, ON, Canada\\
$\ddagger$ Department of Artificial Intelligence and Informatics, Mayo Clinic, Rochester, MN, USA\\{\tt\small tizhoosh.hamid@mayo.edu}}
\maketitle
\begin{abstract}
Learning suitable Whole slide images (WSIs) representations for efficient retrieval systems is a non-trivial task. The WSI embeddings obtained from current methods are in Euclidean space not ideal for efficient WSI retrieval. Furthermore, most of the current methods require high GPU memory due to the simultaneous processing of multiple sets of patches. To address these challenges, we propose a novel framework for learning binary and sparse WSI representations utilizing a deep generative modelling and the \emph{Fisher Vector}. We introduce new loss functions for learning sparse and binary permutation-invariant WSI representations that employ instance-based training achieving better memory efficiency. The learned WSI representations are validated on The Cancer Genomic Atlas (TCGA) and Liver-Kidney-Stomach (LKS) datasets. The proposed method outperforms \emph{Yottixel} (a recent search engine for histopathology images) both in terms of retrieval accuracy and speed. Further, we achieve competitive performance against SOTA on the public benchmark LKS dataset for WSI classification.

\end{abstract}

\section{Introduction}
\label{sec:intro}

The widespread adoption of digital pathology has spurred the digitization of tissue biopsy samples, known as whole slide images (WSIs) ~\cite{bera2019artificial}. The computational pathology is expected to reduce the physicians' workload, improve diagnostic performance, and facilitate the teaching and research in pathology~\cite{tizhoosh2018artificial}. Deep learning is a successful tool for image analysis, including various applications in the medical domain. However, deep networks are challenging to adapt for WSI analysis~\cite{dimitriou2019deep}. These challenges include, but not limited to---
tissue textures, rotationally invariant nature of the tissue, staining variations, and lack of fine-grained (patch-level) labelled data. Among all challenges, the major challenge is the sheer size of WSIs, typically $\gg$ 50,000 $\times$ 50,000 pixels. Furthermore, WSIs are arranged in a multi-resolution pyramidal structure containing images at different magnifications~\cite{tizhoosh2018artificial}. Therefore, memory and computationally efficient frameworks for WSI analyzing are an urgent need \cite{niazi2019digital}.

Patch extraction is typically the first step for the representation learning of a WSI. Commonly, thousands of representative patches can be extracted from a WSI. Processing the patches separately instead of the entire WSI eases the memory bottleneck; however, this leads to multi-vector embedding, which is non-trivial to transform to a single vector representation introducing new challenges, e.g., high data usage and compromised retrieval speed \cite{kalra2020yottixel}. Computing a single-vector representation of a WSI is an active area of research~\cite{hemati2021cnn, tellez2019neural,kalra2020learning}. Ideally, we are interested in a deep-learning solution that can be efficiently trained on WSI patches (at various magnifications), yielding a compact single-vector representation for the WSI, much more suitable for efficient retrieval tasks.

Multiple instance learning (MIL) enables learning on set data instead of using single instances during training. MIL is an appropriate method applicable to WSI representation, and as a result, there is a large body of papers exploring various MIL schemes for WSI representation learning ~\cite{dietterich1997solving,hemati2021cnn,quellec2017multiple,hou2016patch,ilse2018attention,kalra2020learning}. Although MIL has become a preferred method for WSI representation, it does have several limitations. First, the obtained euclidean embeddings cannot be directly used for WSI search in its raw form. Searching within large archives of WSIs through the nearest neighbour search would lead to a prohibitively large increase in memory demand and retrieval times~\cite{7915742}.  As a result, the ancillary processing method is usually necessary to encode these embeddings into more suitable forms, i.e., binary and sparse embeddings facilitating the speed and memory efficiency in nearest neighbour search. Second, current MIL-based methods require all instances to be processed at once as a set (called \emph{bag}), making it difficult to develop end-to-end training in a memory-efficient manner. Finally, current WSI engines,  i.e., \emph{Yottixel} \cite{kalra2020yottixel}, \emph{SMILY}~\cite{hegde2019similar} ignore the a-priori knowledge, such as tumor type, about WSIs for performing the search. It is ideal to employ all known attributes of WSIs for producing a more effective embedding. For this research, our contributions are as follows: 1) The compact (sparse and binary) and permutation-invariant WSI representations ideal for efficient WSI search in large archives, 2)  the permutation-invariant representation of WSIs, trained end-to-end by feeding individual instances instead of a bag of instances which eases up the time and memory bottlenecks, enabling our methods to even incorporate patches at multiple magnification levels, 3)  Learning representations guided by a-priori information, i.e., the tumor type as a way of self-supervision.


The rest of this paper is organized as follows: In \cref{sec:related} we briefly review the current related works on WSI representation learning. Then, in \cref{sec:method} we provide the details of our proposed framework. Next, in \cref{sec:res} we validate the effectiveness of our approach for search and classification tasks on two publicly available benchmark datasets. Finally, we conclude the paper in the \cref{sec:conclusion}.

\section{Related works}
\label{sec:related}
In this section, we review the related literature on WSI representation learning. We organized the related literature into three main themes, i.e., heuristic deep architectures, Muti-instance Learning (MIL)-based methods, and dictionary learning approaches.

\noindent \textbf{Heuristic architectures.}
These methods generally split the task into multiple separate steps to simplify the problem. First, there is an instance-based training where instances are smaller parts of WSIs, typically patches. Then, another network is trained to obtain WSI embeddings while capturing the spatial relationship between patches. For example, Bejnordi et al.~\cite{bejnordi2017context} proposed to employ two sub-networks for processing high and low-resolution information separately and then attaching two networks together. Other works in this category are Spatio-Net~\cite{kong2017cancer} and the neural compression scheme proposed by Tellez et al.~\cite{tellez2019neural}. In Spatio-Net, a grid of embeddings for each patch and its neighbours are obtained by a CNN feature extractor, and then they are processed by  2D-LSTM layers to capture the spatial information. Tellez et al.~\cite{tellez2019neural} proposed a two-stage neural compression where the first stage is devoted to unsupervised representation learning of grid of all image patches per WSI. Then, they employed this trained model to obtain compressed patches and WSI. Finally, in one recent work, authors in \cite{maksoud2020sos} proposed a framework to choose between low and high-resolution information for WSI classification.


\noindent \textbf{Multiple instance learning (MIL).}
Representing each WSI as a bag of patches makes MIL-based schemes a natural approach for end-to-end WSI representation learning~\cite{dietterich1997solving,quellec2017multiple,kalra2020learning,ilse2020deep,ilse2018attention}. One of the early works in MIL-based WSI classification was conducted by Hou et al.~\cite{hou2016patch} where they first trained a patch level classifier and then a fusion model using MIL scheme to achieve WSI classification. In fact, one can regard this approach as a two-step instance-based MIL method where an algorithm determines instance classes. Motivated by this, Chikontwe et al.~\cite{10.1007/978-3-030-59722-1_50} proposed an end-to-end MIL-based method for simultaneous patch and WSI representation learning in a single framework where a center loss is introduced to map patch embeddings from the same WSI to a single centroid. Their approach achieved promising results compared with other MIL-based methods, especially two-stage MIL methods. Other recent MIL methods include \cite{ilse2018attention} and \cite{hemati2021cnn} where pooling layers based on attention mechanism and Deep Sets \cite{zaheer2017deep} have been proposed. Finally, Kalra et al.~\cite{kalra2021pay} employed focal factor learning to modulate the aggregated patch-level predictions. 

\noindent \textbf{Dictionary learning.}
Another approach that can be used for the WSI representation is the \emph{bag of visual words} (BoVW)~\cite{csurka2004visual}  for encoding local image descriptors into one embedding. A more advanced version of BoVW that captures higher-order statistics to obtain the set representation is based on the \textit{Fisher Kernel} theory and generative models~\cite{jaakkola1999exploiting}. Authors in~\cite{perronnin2007fisher} introduced  Gaussian mixture model (GMM)-based Fisher Vector which can be calculated using the normalized gradient of the log-likelihood of the GMMs with respect to parameters, such as mixing coefficients, means, and variances, given a set of observations. Further, recently there has been some research to extract Fisher Vector from deep generative models \cite{qiu2017deep,zhai2019adversarial}. Although this set encoding ability makes Fisher Kernel a natural candidate for WSI representation learning. There are only a few papers that use Fisher theory and dictionary learning in general for the WSI representation task~\cite{song2017supervised, song2017adapting, zhu2018multiple}. The reason could be attributed to the fact that Fisher Vector  is formulated in a fully unsupervised manner using GMMs. However, considering the challenges inherent to pathology images (e.g., complex textures and colour variations), employing available WSI information, i.e., tumor type and primary diagnosis, in obtaining an efficient global representation is necessary. Further, GMM-based Fisher Vector captures no more than second-order statistics of data for set encoding. Besides, the training of GMMs is sub-optimal and not end-to-end. Finally, the obtained encodings are generally high-dimensional embeddings in Euclidean space, which are less desirable for WSI search due to their increased computation times for the distance computation.
\section{Method}
\label{sec:method}
This section presents the proposed framework for learning compact WSI representations. First, we briefly review the relevant concepts, i.e., Fisher Kernel~\cite{jaakkola1999exploiting} and Fisher Vector theories~\cite{perronnin2007fisher}. Next, we describe the proposed method based upon variational autoencoders (VAEs) and Fisher Vector theory. The proposed method is memory efficient during training and learns representations that are permutation-invariant, compact (sparse/binary), and can be conditioned on known information (e.g., the given tumor type) for the self-supervision. The proposed method is trained in an end-to-end manner on individual instances instead of a bag of instances to obtain representations for both patches and the WSI in its entirety.

\subsection{Preparation}
The key idea behind Fisher Kernel is to derive the kernel function from a generative probability model. Initially, the main motivation for deriving such kernels was bridging the gap between generative and discriminative models~\cite{jaakkola1999exploiting}: \say{the gradient of the log-likelihood with respect to a parameter describes how that parameter contributes to the process of generating a particular example}. As a result, to take advantage of generative models in discriminative tasks, Jaakkola and Haussler proposed to employ the gradient space of the generative models to use the generative process as a similarity metric between examples (or set of examples, i.e., $\mathbf{X}=\{\mathbf{x}_t,t=1,\dots, T\}$ where $T$ is the number of examples in the set) \cite{jaakkola1999exploiting}. Let us consider a class of probability models $p(\mathbf{X} \mid \pmb{\theta})$ where $\pmb{\theta}  \in  \pmb{\Theta}$ is a parameter vector and $\mathbf{X}$ is set of examples, i.e., $\mathbf{X}=\{\mathbf{x}_t,t=1,\dots, T\}$. The Fisher Score is then defined as
\begin{align}
U_\mathbf{X}= \nabla_{\pmb{\theta}} \log p(\mathbf{X} \mid \mathbf{\pmb{\theta}}),
\label{eq:1}
\end{align} 
where the $U_\mathbf{X} \in \mathbb{R}^d$. The dimensionality $d$ of the Fisher Score is equal to the number of parameters in the generative model $p(\mathbf{X} \mid \pmb{\theta})$ independent of the number of data points in the set $T$. The Fisher information matrix (FIM) is 
\begin{align}
\mathbf{I}=E_{\mathbf{x}\sim p(\mathbf{x} \mid \pmb{\theta})}\{U_\mathbf{X}  U_\mathbf{X}^T\}.
\label{eq:2}
\end{align} 
Subsequently, the Fisher Kernel can be defined as
\begin{align}
K(\mathbf{X},\mathbf{Y})=U_\mathbf{X}^{T} \mathbf{I}^{-1} U_\mathbf{Y}
\label{eq:3}
\end{align} 

Fisher Kernel can be used to calculate the similarity between two sets of data points~\cite{jaakkola1999exploiting}. Authors in~\cite{perronnin2007fisher} proposed the GMM-based Fisher Vector as a way to encode a set of local descriptors in a single embedding where the Fisher Vector is the normalized Fisher Score ($\mathbf{s}_F$) calculated as 
\begin{equation}
\begin{aligned}
\mathbf{s}_F\!=\!\frac{1}{T} \mathbf{L} \nabla_{\pmb{\theta}} \log p(\mathbf{X} \mid \pmb{\theta})\!=\!\mathbf{L} \frac{1}{T}   \sum_{t=1}^{T}  \nabla_{\pmb{\theta}} \log p(\mathbf{x}_t \mid \pmb{\theta}),
\label{eq:4}
\end{aligned}
\end{equation}
where $\mathbf{L}$ is calculated from Cholesky decomposition of inverse FIM, i.e., $\mathbf{I}^{-1}= \mathbf{L}^T \mathbf{L}$, with that assumption that data points in $\mathbf{X}$  are statistically independent.



\begin{figure*}[ht]
  \centering
  \centering\includegraphics[width=.7
  \linewidth]{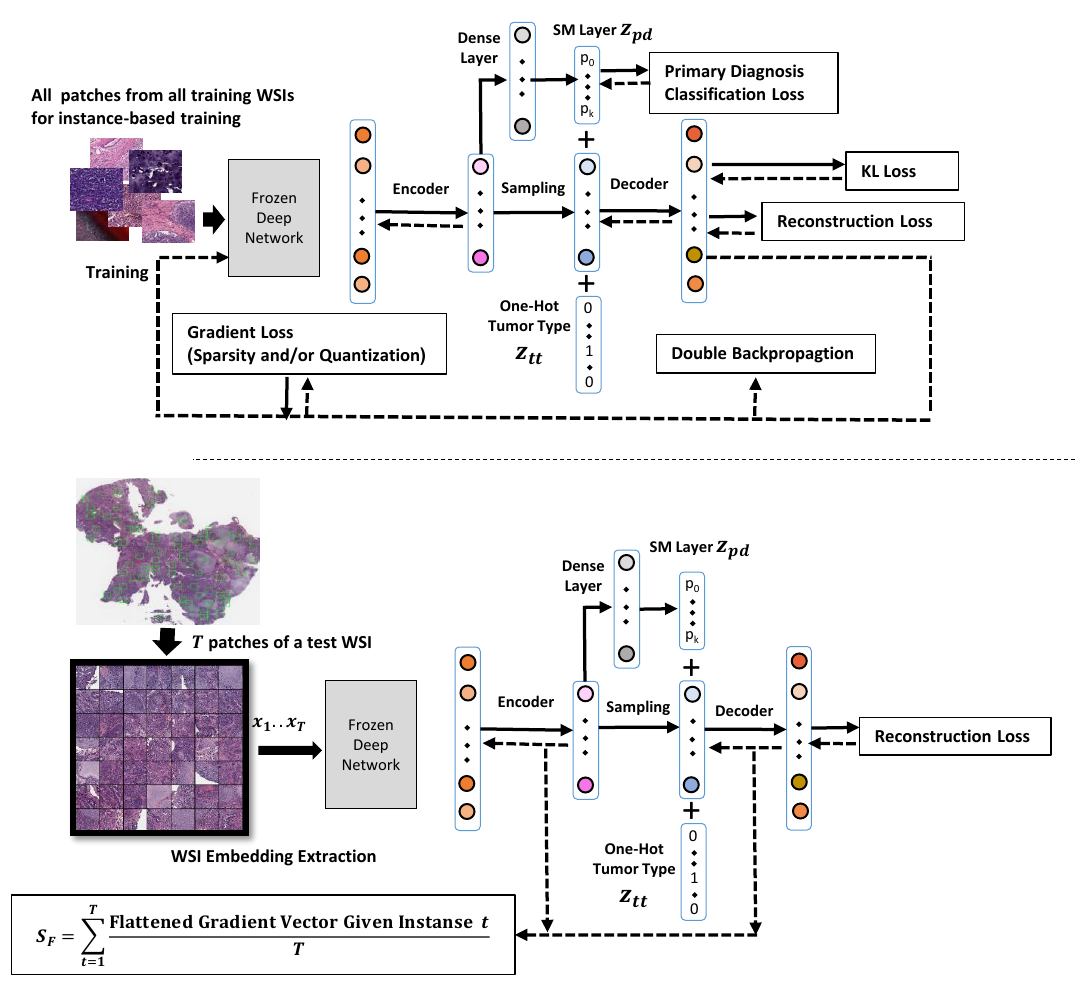}
   \caption{The first row represents the proposed architecture and associated instance-based training scheme. The second row shows the procedure for obtaining the WSI embedding for a set of patches of this WSI given the trained model.}
    \label{fig:1}%
  \end{figure*}

\subsection{Deep Compact Fisher Vector}

The GMM-based Feature Vector exhibits shortcomings such as lack of employing available information, ignoring higher-order statistics in set encoding, its sub-optimal optimization, non-end-to-end training scheme, and the fact that Euclidean embeddings are not feasible for large archives. Motivated to remove these limitations, we propose a new type of Fisher Vector based on the deep generative models for the WSI representation learning. The contributions of our method are as follows.

\begin{enumerate}
  \item To capture higher-order statistics while learning the set representation, we propose to employ generative models  VAE here ~\cite{kingma2019introduction} for WSI representation learning.
  \item We add a classification loss to the training such that the available WSI level primary diagnosis labels are employed during the training of the VAE.
  \item We design the VAE to be conditioned on available information, e.g., the tumor type. Given the fact that every tumor type has its own specific cancer subtypes, this conditioning is expected to improve the quality of WSI embeddings.
  \item More importantly, we propose two novel loss functions for compact (sparse and binary) and permutation-invariant WSI representation learning.
\end{enumerate}

We start by training a VAE and then modify the VAE to be conditioned on tumor type. We add a classification loss to the end of the encoder part such that primary diagnosis label information is injected into the model space. Finally, we propose two novel loss functions for learning sparse and binary permutation-invariant representations.
    

\noindent \textbf{VAE loss function --}
To learn the encoder and decoder parameters of the VAE, i.e., $\pmb{\phi}$ and $\pmb{\theta}$, that models distribution of $\mathbf{x}$, we assume the prior distribution on the random variable $\mathbf{z}$ is $p_{\mathbf{\pmb{\theta}}}(\mathbf{z})$ and as a result, $\mathbf{x}_t$ is sampled from $p_{\pmb{\theta}}(\mathbf{x} \mid \mathbf{z})$. In this case, one can show that the lower bound for the  $\log p_{\pmb{\theta}}(\mathbf{x})$ can be calculated as
\begin{equation}
\log p_{\pmb{\theta}}(\mathbf{x})\!\geq\! - q_{\pmb{\phi}}(\mathbf{z} \mid \mathbf{x}){p_{\pmb{\theta}}(\mathbf{z})}\!+\! E_{q_{\pmb{\phi}}(\mathbf{z} \mid \mathbf{x})}\left[\log p_{\pmb{\theta}}(\mathbf{x} \!\mid\! \mathbf{z})\right],
\label{eq:5}
\end{equation} 
where $q_{\pmb{\phi}}(\mathbf{z} \mid \mathbf{x})$ is the  approximate posterior with parameters $\pmb{\phi}$. The lower bound in Eq. \ref{eq:5} is known as variational lower bound and on the patch $\mathbf{x}_t$ and it is represented with $\mathcal{LB}(\pmb{\phi},\pmb{\theta},\mathbf{x}_t)$. We aim to maximize $\mathcal{LB}$ to learn the generative model parameters. In context of the VAE model,  $q_{\pmb{\phi}}(\mathbf{z}_t \mid \mathbf{x}_t)$ and  $p_{\pmb{\theta}}(\mathbf{x}_t\mid\mathbf{z}_t)$ are encoder and decoder, respectively. In order to learn the encoder and decoder parameters, i.e., $\pmb{\phi}$ and $\pmb{\theta}$, first we assume the prior distribution $p_{\pmb{\theta}}(\mathbf{z}_t)$ is   $\mathcal{N}(\mathbf{z}_t;0,\mathbf{I})$ and $q_{\pmb{\phi}}(\mathbf{z}_t \mid \mathbf{x}_t)$ and $p_{\pmb{\theta}}(\mathbf{x}_t\mid\mathbf{z}_t)$ follow the normal distributions $\mathcal{N}(\mathbf{z}_t;\pmb{\mu}_{\mathbf{z}_t},\pmb{\sigma}^2_{\mathbf{z}_t}\mathbf{I})$ and $\mathcal{N}(\mathbf{x}_t;\pmb{\mu}_{\mathbf{x}_t},\pmb{\sigma}^2_{\mathbf{x}_t}\mathbf{I})$. In this case, by estimating the latent code using one-step Monte Carlo, the variational lower bound is
\begin{equation}
\begin{aligned}
\mathcal{LB}(\pmb{\phi},\pmb{\theta},\mathbf{x}_t)\!=\! \log p_{\pmb{\theta}}(\mathbf{x}_t  \mid \mathbf{z}_t)\!+\! \frac{1}{2} \sum_{j=1}^{d}(1+\log \pmb{\sigma}^2_{\mathbf{z}_t(j)})\\
-\frac{1}{2} \lVert \pmb{\mu}_{\mathbf{z}_t} \rVert ^2 -\!\frac{1}{2} \lVert \pmb{\sigma}_{\mathbf{z}_t} \rVert ^2,
\label{eq:6}
\end{aligned} 
\end{equation} 
where $\mathbf{z}_t$ is sampled from $\mathcal{N}(\pmb{\mu}_{\mathbf{z}_t},\pmb{\sigma}^2_{\mathbf{z}_t}\mathbf{I})$.

\noindent \textbf{Conditioned VAE --} As the tumor type of a WSI is always available, we conditioned VAE on the tumor type of the given WSI to draw benefit from this apriori knowledge. Following the~\cite{sohn2015learning}, let's represent the tumor type as a one-hot encoded vector $\mathbf{z}_{tt}$, then we concatenate this vector to the $\mathbf{z}_t$. Furthermore, to inject WSI-level primary diagnosis information into the generative model, we add a classification loss to the last layer of the encoder before the sampling layer. We assign the WSI label to all patches extracted from that WSI. Then, we concatenate the softmax of predicted primary diagnosis $\mathbf{z}_{pd}$, with length $k$, to the latent space. The latent space associated with $\mathbf{x}_{t}$ that is fed to decoder is modified as   $\mathbf{z}_{t} \leftarrow [\mathbf{z}_{t},\mathbf{z}_{tt},\mathbf{z}_{pd}]$. Considering the classification loss, so far, the loss function for training the conditioned VAE (CVAE) has the form
\begin{equation}
\mathbf{\mathcal{L}}_\textrm{CVAE}=\lambda_1 \mathbf{\mathcal{L}}_{rec} + \lambda_2 \mathbf{\mathcal{L}}_{kl}+ \lambda_3 \mathbf{\mathcal{L}}_{cls},
\label{eq:7}
\end{equation} 
where minimizing the first two terms is equivalent to maximizing the variational lower bound, and the third loss is the classification loss of predicting cancer subtypes.

\noindent \textbf{Deep Sparse Fisher Vector --}
Now, we propose a novel method for learning Sparse Fisher Vector (SFV). As the gradient space represents the WSI, we encourage sparsity in the gradient  by adding the $l_1$ norm of the gradient of the loss function in Eq.\ref{eq:7} to the overall training loss. To regularize the gradient, we utilize the \emph{double backpropagation}, where given a batch of data points $\mathbf{X}$; the loss function can be written as
\begin{equation}
\mathbf{\mathcal{L}}_\textrm{SFV}=\mathbf{\mathcal{L}}_\textrm{CVAE} + \lambda_4   \sum_{\mathbf{W}_i \in \mathbb{W}} \|\nabla_{\mathbf{W}_i} \mathbf{\mathcal{L}}_\textrm{CVAE}(\mathbb{W},\mathbf{X})\|_1,
\label{eq:8}
\end{equation}
where $\mathbb{W}$ is the set of CVAE parameters for all layers, $\mathbf{W}_i$  and $\nabla_{\mathbf{W}_i}\mathbf{\mathcal{L}}_\textrm{CVAE}(\mathbb{W},\mathbf{X})$ are the is parameters and the gradient of the CVAE loss with respect to the $i^{th}$ layer parameters. To the best of our knowledge, such an end-to-end Sparse Fisher Vector learning does not exist in the literature. 

\noindent \textbf{Deep Binary Fisher Vector --}
For learning deep binary Fisher Vector (BFV), inspired by the quantization-based learning in hashing literature~\cite{gong2012iterative,hemati2020non}, we propose to reduce the quantization loss of the gradient of the CVAE loss with respect to each layer's parameters. We propose to find 
\begin{equation}
\begin{aligned}
& \underset{\mathbf{B}_i,\nabla_{\mathbf{W}_i}  \mathbf{\mathcal{L}}_\textrm{CVAE}(\mathbb{W},\mathbf{X})}{\text{arg min}} &&  \sum_{\mathbf{W}_i \in \mathbb{W}} \|\nabla_{\mathbf{W}_i} \mathbf{\mathcal{L}}_\textrm{CVAE}(\mathbb{W},\mathbf{X})\!-\!\mathbf{B}_i \| ^2_2 
\\ 
& \textrm{s.t.}
&& \mathbf{B}_i \in \{-1,1\} ^{d_i\times 1},
\label{eq:9}
\end{aligned}
\end{equation}
 where $\mathbf{B}_i$ is the flattened binary representation of the gradient of the CVAE loss w.r.t parameters of the $i^{th}$ layer. In this case, the loss function to obtain BFV can be written as
\begin{align}
\mathbf{\mathcal{L}}_\textrm{BFV}\!=\!\mathbf{\mathcal{L}}_\textrm{CVAE}\!+\! \lambda_5   \sum_{\mathbf{W_i} \in \mathbb{W},\mathbf{B_i} \in \mathbb{B}} \! \|\nabla_{\mathbf{W}_i} \mathbf{\mathcal{L}}_\textrm{CVAE}(\mathbb{W},\mathbf{X})\!-\!\mathbf{B}_i \| ^2_2 ,
\label{eq:10}
\end{align}
 where $\mathbb{B}$ is the set of closest hamming vertices to gradients w.r.t all layers. Given the binary optimization variable $\mathbf{B}_i$, on each epoch, we employ the coordinate descent approach and update each of $\mathbf{B}_i$ and  $\mathbf{W}_i$ while the other is fixed. 
For the case that $\mathbf{W}_i$ is fixed the problem turns to

\begin{equation}
\begin{aligned}
& \underset{\mathbf{B}_i}{\text{arg min}} && \|\nabla_{\mathbf{W}_i} \mathbf{\mathcal{L}}_\textrm{CVAE}(\mathbb{W},\mathbf{X})-\mathbf{B}_i \| ^2_2 
\\ 
& \textrm{s.t.}
&& \mathbf{B}_i \in \{-1,1\} ^{d_i\times 1},
\label{eq:11}
\end{aligned}
\end{equation}
where by expanding Eq.\ref{eq:11} it turns out the above minimization is equivalent to

\begin{equation}
\begin{aligned}
& \underset{\mathbf{B}_i}{\text{arg max}} && \mathbf{B}_i^T. \nabla_{\mathbf{W}_i} \mathbf{\mathcal{L}}_\textrm{CVAE}(\mathbb{W},\mathbf{X}) 
\\ 
& \textrm{s.t.}
&& \mathbf{B}_i \in \{-1,1\} ^{d_i\times 1}.
\label{eq:12}
\end{aligned}
\end{equation}

This problem has the following closed-form solution \cite{gong2012iterative}:
\begin{equation}
\mathbf{B}_i=sgn(\nabla_{\mathbf{W}_i} \mathbf{\mathcal{L}}_\textrm{CVAE}(\mathbb{W},\mathbf{X})).
\label{eq:13}
\end{equation}

The loss function can be for fixed  $\mathbf{B}_i$ as 
\begin{align}
\mathbf{\mathcal{L}}_\textrm{BFV}\!=\!\mathbf{\mathcal{L}}_\textrm{CVAE}\!+\!\lambda_5\!   \sum_{\mathbf{W_i} \in \mathbb{W}}\! \|\nabla_{\mathbf{W_i}} \mathbf{\mathcal{L}}_\textrm{CVAE}(\mathbb{W},\mathbf{X})\!-\!\mathbf{B}_i \| ^2_2 
\label{eq:14}
\end{align}

This is similar to SFV learning in Eq. \ref{eq:8}. The variables can be updated using double backpropagation.

\noindent \textbf{Deep Sparse Binary Fisher Vector --}
Knowing that the length of obtained WSI embeddings is equal to the number of parameters in the generative model, we may be interested in compact (short) binary codes for more efficient WSI retrieval. We propose to employ both gradient sparsity and gradient quantization losses to achieve \emph{Conditioned Sparse Binary Fisher Vector} (C-Deep-SBFV). Gradient sparsity pushes the generative model to use fewer parameters to generate a data point. As a result, the quality of embedding will be more robust to dropping some dimensions, i.e., gradient w.r.t some parameters of VAE. To choose effective dimensions for each tumor type we find the top $M$ parameters that provide the highest variance in their respective gradient values for the training data.

\noindent \textbf{VAE Architecture and Training Scheme --}
The architecture of the proposed conditioned VAE is given in the first half of \cref{fig:1}. We employed a frozen pre-trained CNN (DenseNet-121~\cite{huang2017densely}) as the backbone of the VAE. Each encoder and decoder parts contain three fully connected layers. The last layer of the encoder is fed to a softmax layer (SM Layer in \cref{fig:1}) for primary diagnosis prediction. In order to condition the VAE, for each patch, the output of the softmax layer along with a one-hot encoded vector representing the available tumor type information of the patch is concatenated to the latent vector to create the $\mathbf{Z}_t$. Then, this vector is fed to the decoder part. As it can be seen from the \cref{fig:1}, the CVAE is trained on a per-instance basis enabling to include even patches from multiple magnifications.

\noindent \textbf{WSI Embedding Extraction --} After the training phase, to obtain a single embedding for a WSI, all patches of that WSI are fed to the CVAE (see the second half of \cref{fig:1}). Then, given the reconstruction loss, we calculate the average gradient over all patches using backpropagation to obtain the Fisher Score ($\mathbf{s}_F$). Based on Fisher Theory, we also need $\mathbf{L}$ obtained from FIM to normalize the vector and derive the Fisher Vector. However,  given the computational load of calculating $\mathbf{L}$, we replace this with identity matrix and normalize the gradient using power and $l_2$ normalization steps proposed by~\cite{perronnin2010improving}. In other words, representing the power and  $l_2$  normalization steps as $\mathcal{S}(\cdot)$ operator,  the conditioned deep compact Fisher Vector $\mathbf{v}_F$ is calculated from the Fisher Score $\mathbf{s}_F$:
\begin{equation}
\begin{aligned}
\mathbf{v}_F\!=\!\mathcal{S}\left(\frac{1}{T}   \sum_{t=1}^{T}  \nabla_{\pmb{\theta}, \pmb{\phi} } \|\mathbf{x}_t -\mathbf{\hat{x}}_t(\pmb{\theta}, \pmb{\phi}) \| ^2_2\right)\! =\!\mathcal{S}\left(\mathbf{s}_F\right).
\label{eq:15}
\end{aligned} 
\end{equation} 
where $\mathbf{x}_t$ and $\mathbf{\hat{x}}_t(\pmb{\theta}, \pmb{\phi})$ are the patch embedding and its reconstruction. The size of the proposed feature vector is equal to the number of parameters in CVAE. The test-time, the one-hot vector of the tumor type, will be fed to the CVAE as a known parameter while the $\mathbf{z}_{pd}$ is calculated by the classifier.

\section{Results}
\label{sec:res}
We evaluate the quality of the WSI embeddings obtained by the proposed method for both search and classification tasks. The datasets we employed are diagnostic slides from The Cancer Genomic Atlas (TCGA) repository ~\cite{weinstein2013cancer} and the Liver-Kidney-Stomach (LKS) immunofluorescence  ~\cite{maksoud2020sos} to conduct experiments.
\begin{table*}[ht]
\centering
{\scriptsize
\begin{tabular}{llrcccccc}
\toprule
       
Site & Subtype & $n_\textrm{slides}$ & Yottixel & C-GMM-FV & C-Deep-FV  & C-Deep-SFV  &C-Deep-BFV  &C-Deep-SBFV   \  \\ 
\toprule

\multirow{2}{*}{Brain}                 & LGG &323                    &86.60                         &85.35  &92.83  &93.07                          &93.10                      &{\cellcolor[rgb]{.749,1,  .749}}93.43  \\
                 &GBM  &387                     &88.68                         &87.63     &93.44
                &93.76                         &{\cellcolor[rgb]{.749,1,  .749}}93.99        &94.24\\
\hline

\multirow{3}{*}{Endocrine}                 
& THCA &198                    &97.98                         &97.02    &98.74    &{\cellcolor[rgb]{.749,1,  .749}}99.24                        &{\cellcolor[rgb]{.749,1,  .749}}99.24               &98.50       \\

& ACC  & 93                     &93.68                         &91.01  &94.62  &95.08                   &94.62       &{\cellcolor[rgb]{.749,1,  .749}}95.13           \\
                 
& PCPG & 70                   &92.53                         &87.14   &91.97          &{\cellcolor[rgb]{.749,1,  .749}}92.95                         &90.64         &91.97           \\ 
\hline
\multirow{4}{*}{Gastro.} 
& ESCA & 55                    &60.95
                           &58.82           &72.00    &65.42                           &{\cellcolor[rgb]{.749,1,  .749}}75.43     &66.66   \\

& COAD & 174                    &72.62                          &71.95  &72.72 &74.93             &74.44           &{\cellcolor[rgb]{.749,1,  .749}}76.42  \\

& STAD &157                    &79.75                          & 79.75   &78.59 &80.75                          &83.48        &{\cellcolor[rgb]{.749,1,  .749}}83.97          \\

& READ & 61                    & 24.24                       &29.62  &23.52 &{\cellcolor[rgb]{.749,1,  .749}}31.77                         & 30.30     &31.37    \\ 
\hline
\multirow{3}{*}{Gynaeco.}           

& UCS  &37                     &65.62                        &66.66  &{\cellcolor[rgb]{.749,1,  .749}}78.26    &72.13                          &74.62   &73.01     \\

& UCEC  & 206                     &84.23                          &82.82   &{\cellcolor[rgb]{.749,1,  .749}}89.31 &87.29                     &83.33    &88.16\\

& CESC   & 113                    &71.71                        &76.10    &{\cellcolor[rgb]{.749,1,  .749}}86.36    &81.44                          &70.47  &81.65        \\

& OV   &42                    & 64.78                          &68.42      &76.74   &{\cellcolor[rgb]{.749,1,  .749}}83.95                         &77.33       &80.95     \\ 
\hline

\multirow{2}{*}{Haematopoietic.}            & THYM  &80                     &93.41                          &93.49     &93.56      &93.97                          &{\cellcolor[rgb]{.749,1,  .749}}96.93       &94.04    \\

& DLBC &14                    &47.61                          &42.10  &35.29    &54.54                          &{\cellcolor[rgb]{.749,1,  .749}}80.00       &50.00\\
\hline

\multirow{3}{*}{Liver, panc.}    & CHOL     &17                     &32.00                         &38.46     &40.00     &{\cellcolor[rgb]{.749,1,  .749}}48.27                         &41.66       &32.00\\

& LIHC & 146                    &94.31                          &93.55  &92.61    &93.91                          &94.00         &{\cellcolor[rgb]{.749,1,  .749}}94.38                                        \\

& PAAD & 65                    &93.93
\                           &91.85     &92.18    &{\cellcolor[rgb]{.749,1,  .749}}94.65\                           &93.93                 &93.75    \\ 

\hline

\multirow{2}{*}{Melanocytic malignancies} & SKCM &184                     &96.08                       &98.37    &98.11    &98.37                         &{\cellcolor[rgb]{.749,1,  .749}}98.92       &{\cellcolor[rgb]{.749,1,  .749}}98.92\\

& UVM & 40                    &76.92                          &92.30     &90.90     &92.30                        &{\cellcolor[rgb]{.749,1,  .749}}94.73     &{\cellcolor[rgb]{.749,1,  .749}}94.73                                      \\

\hline

\multirow{2}{*}{Prostate/testis}           & PRAD & 176                    &98.56                          &98.00     &{\cellcolor[rgb]{.749,1,  .749}}99.42   &98.55                       &99.14       &99.14
\\

& TGCT & 112                    &97.79                      &96.88   &{\cellcolor[rgb]{.749,1,  .749}}99.11 &97.81                          &98.67     &98.66\\ 
\hline

\multirow{3}{*}{Pulmonary}                 
& LUAD & 218                    &67.44                       &74.77    &77.96           &79.12                          &74.88     &{\cellcolor[rgb]{.749,1,  .749}}79.13           \\

& LUSC & 198                    &67.75                    &70.02  &71.65     &72.95                          &72.04                 &{\cellcolor[rgb]{.749,1,  .749}}76.14 \\

& MESO & 27                     & 7.14                             &43.24      &50.00        &{\cellcolor[rgb]{.749,1,  .749}}51.28                            &40.00      &31.25     \\ 
\hline
\multirow{4}{*}{Urinary tract}             
& BLCA & 193                    &90.41                           &88.26  &92.34   &92.83                           &94.20    &{\cellcolor[rgb]{.749,1,  .749}}95.93\\
             
& KIRC & 195                    &88.88                          &88.26  &91.82 &93.75                         &91.47     &{\cellcolor[rgb]{.749,1,  .749}}93.81                   \\

& KIRP & 142                    &77.73                          &75.53  &82.97 &84.01                          &84.05    &{\cellcolor[rgb]{.749,1,  .749}}84.78\\

& KICH & 47                    &79.06                          &84.78  &86.36 &{\cellcolor[rgb]{.749,1,  .749}}89.58                          &89.36     &87.50\\ 
\bottomrule
\end{tabular}

\caption{F1-measure (in \%) for majority-3 search  through $k$-NN of the vertical search among 3770 test WSIs for Yottixel,  C-GMM-FV,  C-Deep-FV, C-Deep-SFV, C-Deep-BFV, and C-Deep-SBFV. Best F1-measure values highlighted.} 

\label{tab:Table1}
}

\end{table*}

\begin{figure*}[t]
  \centering
    \centering\includegraphics[width=0.7\columnwidth]{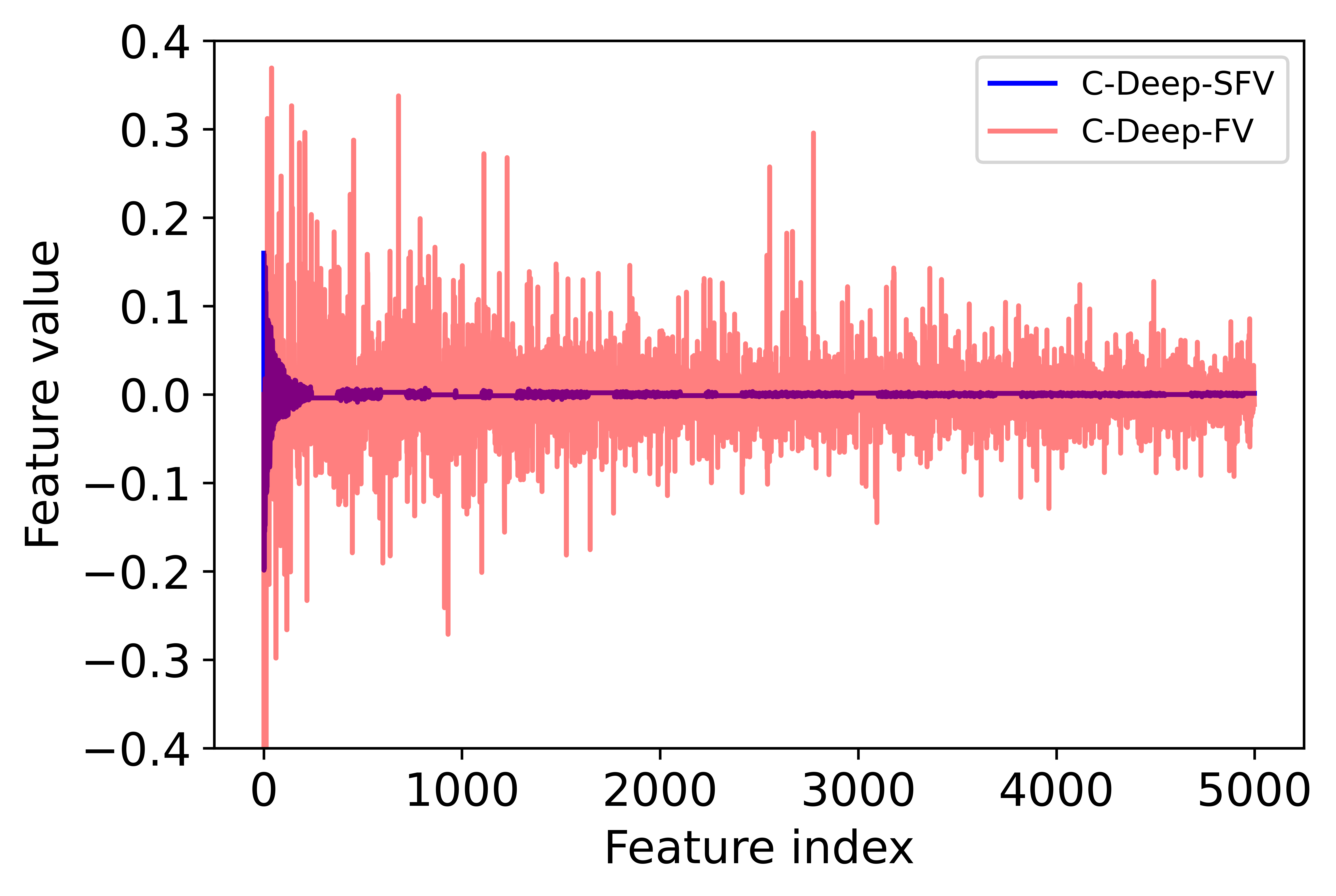}
    \centering\includegraphics[width=0.7\columnwidth]{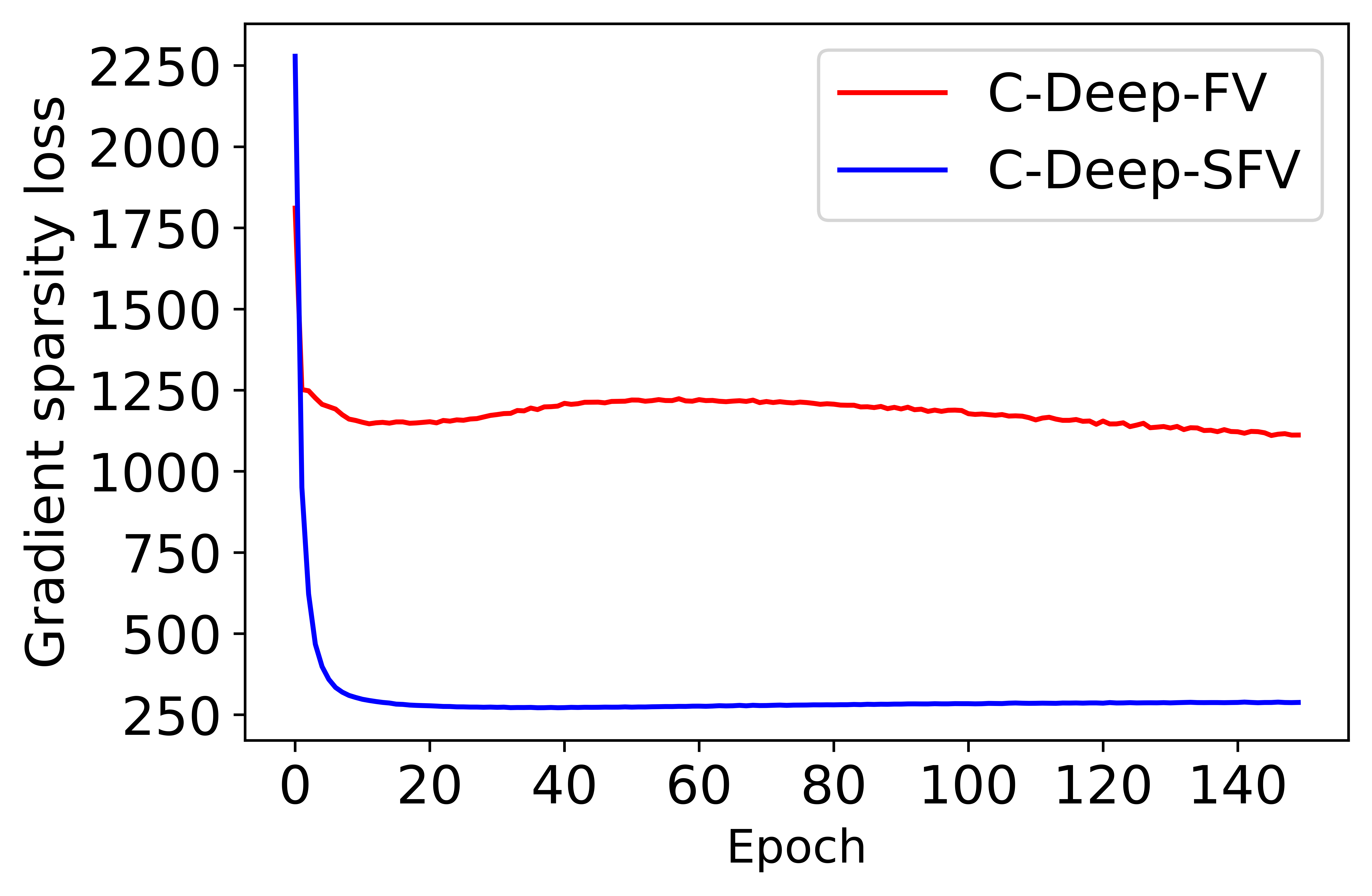}
  \caption{Left: Feature values across first 5,000 high variance dimensions for a WSI using C-Deep-SFV and C-Deep-FV. Right: Gradient sparsity loss ($l_1$ norm of the loss function gradient) of C-Deep-SFV and C-Deep-FV during the training epochs.}%
  \label{fig:2}%
\end{figure*}

\begin{figure*}
  \centering

    
    \centering\includegraphics[width=0.9\columnwidth]{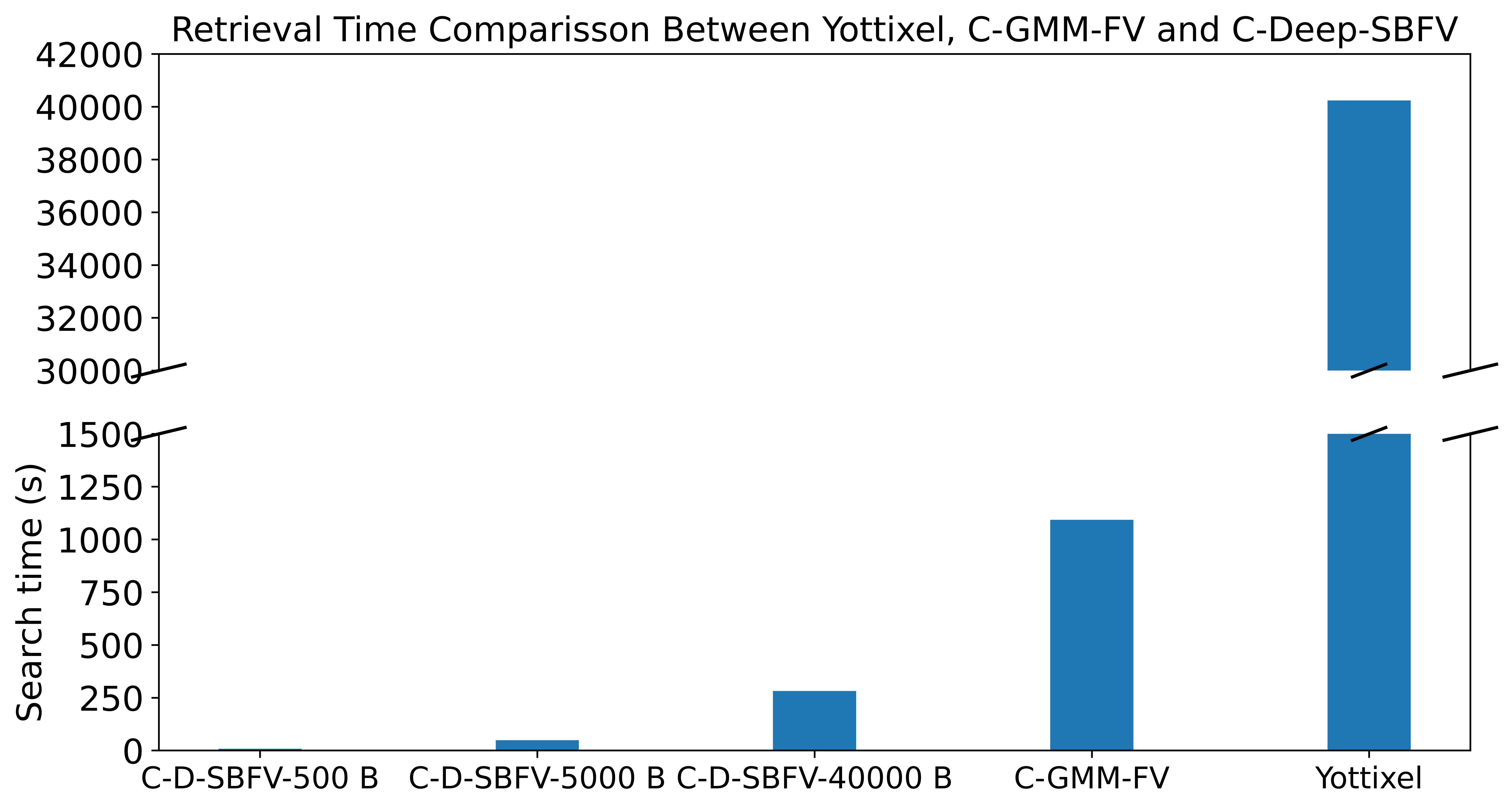}
    \centering\includegraphics[width=0.65\columnwidth]{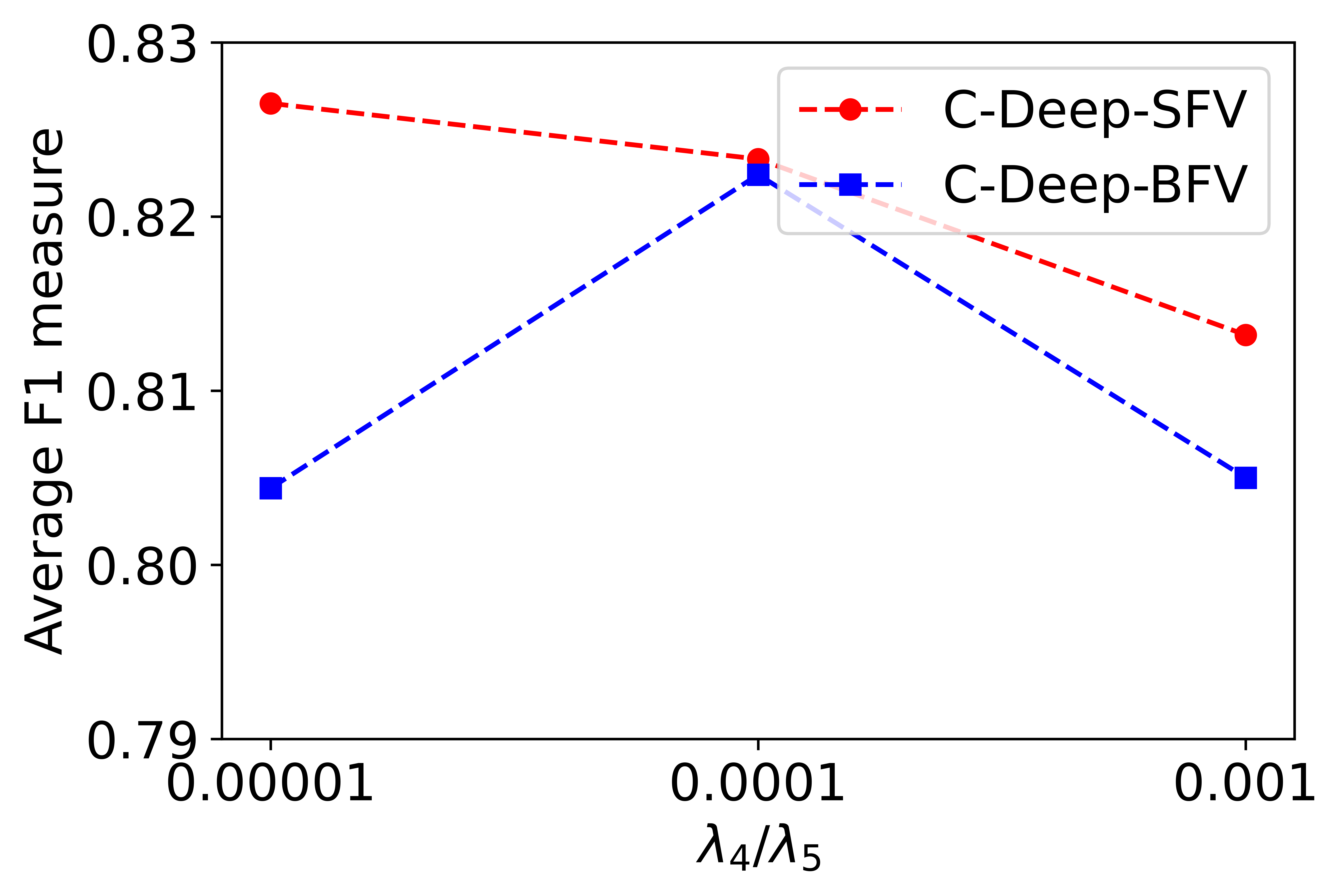}
  \caption{Left: Retrieval times for the leave-one-patient-out search (Tab. \ref{tab:Table1}) for Yottixel and C-Deep-SBFV with different number of bits. Right: ablation study for gradient sparsity and quantization regularization terms. (full results in \cref{tab:Table4}). }%
  \label{fig:3}%
\end{figure*}

\noindent \textbf{WSI search --} For this experiment, we randomly selected $40\%$ of the TCGA diagnostic WSIs as a test set and the rest for training. For both test and training WSIs, $15\%$ of patches with $1000\times1000$ patch size have been selected based on the Yottixel (the same clustering method has been applied)\cite{kalra2020yottixel}. Similar to \cite{kalra2020yottixel}, the vertical search has been applied on the test set (3,761 WSIs), and leave-one-out patient performed for searching WSIs through the same primary site. The majority of the top 3 similar cases have been used for predicting each query cancer subtype. \cref{tab:Table1} compares F1-measure between Yottixel, Conditional GMM-based Fisher Vector (C-GMM-FV), C-Deep-FV, C-Deep-SFV, C-Deep-BFV, and C-Deep-SBFV. Yottixel takes the median of minimum patch distances to calculate two WSIs dissimilarity, while C-GMM-FV and our proposed method obtain one embedding per WSI. Our proposed method improved the search F1-measure for all 29 cancer subtypes while the embeddings are binary and/or sparse. Although in almost all subtypes of two primary sites (Gynecological and Prostate/testis), C-Deep-FV performed better than other methods, almost in all cases, compact WSI embeddings obtained by gradient sparsity and quantization losses achieve even better search performance (see \cref{tab:Table1}). The compactness of the proposed embeddings leads to high efficiency for WSI search in terms of memory usage and retrieval times. \cref{fig:2} (left) shows the embedding for C-Deep-FV and C-Deep-SFV across the first 5,000 high variance dimensions given the tissue type of the given WSI out of 1,407,105 parameters of our CVAE. Considering \cref{fig:2} (a), after encouraging sparsity on the gradients, the C-Deep-SFV can represent the WSI by significantly much fewer parameters leading to compact representations. \cref{fig:2} (right) shows the effectiveness of incorporating gradient sparsity loss in reducing the $l_1$ norm of the loss function gradient during the epochs. 

\begin{table*}[ht]
\centering
{\scriptsize
\begin{tabular}{cccc|cc|cc}
\toprule
  
    &     &   \multicolumn{2}{c}{500 Bits}       &        \multicolumn{2}{c}{5000 Bits}  &        \multicolumn{2}{c}{40000 Bits}      \\\cline{3-4} \cline{5-6} \cline{7-8} 
       
Site & Subtype  & C-Deep-BFV  & C-Deep-SBFV  & C-Deep-BFV  & C-Deep-SBFV & C-Deep-BFV  & C-Deep-SBFV    \\ 
\toprule

\multirow{2}{*}{Brain}                 

& LGG                     &84.48          &{\cellcolor[rgb]{.749,1,  .749}}89.02                                           &86.27         &{\cellcolor[rgb]{.749,1,  .749}}90.93               &93.21       &{\cellcolor[rgb]{.749,1,  .749}}94.06             \\

&GBM          &86.55                         &{\cellcolor[rgb]{.749,1,  .749}}89.83                                       &87.68          &{\cellcolor[rgb]{.749,1,  .749}}92.38      &93.91          &{\cellcolor[rgb]{.749,1,  .749}}94.76 \\
\hline

\multirow{3}{*}{Endocrine}                 
& THCA                   &92.97                         &{\cellcolor[rgb]{.749,1,  .749}}93.62                                    &94.43        &{\cellcolor[rgb]{.749,1,  .749}}95.26         &{\cellcolor[rgb]{.749,1,  .749}}98.48       &98.24        \\

& ACC                       &82.87                         &{\cellcolor[rgb]{.749,1,  .749}}84.37                                                  &86.91    &{\cellcolor[rgb]{.749,1,  .749}}93.19        &94.68       &{\cellcolor[rgb]{.749,1,  .749}}96.25        \\
                 
& PCPG                   &73.43                         &{\cellcolor[rgb]{.749,1,  .749}}75.40                          &69.49       &{\cellcolor[rgb]{.749,1,  .749}}84.61         &91.30       &{\cellcolor[rgb]{.749,1,  .749}}92.64   \\ 
\hline
\multirow{4}{*}{Gastro.}    
& ESCA                     &35.41
                           &{\cellcolor[rgb]{.749,1,  .749}}52.08                                          &41.37     &{\cellcolor[rgb]{.749,1,  .749}}51.02       &{\cellcolor[rgb]{.749,1,  .749}}73.58       &72.89  \\

& COAD                     &64.37                          &{\cellcolor[rgb]{.749,1,  .749}}70.96                       
           &68.16      &{\cellcolor[rgb]{.749,1,  .749}}70.96      &{\cellcolor[rgb]{.749,1,  .749}}77.65       &74.58 \\

& STAD                     &60.81                          &{\cellcolor[rgb]{.749,1,  .749}}78.01   
                         &71.83       &{\cellcolor[rgb]{.749,1,  .749}}75.54       
                         &84.01                &{\cellcolor[rgb]{.749,1,  .749}}85.80          \\

& READ                     &{\cellcolor[rgb]{.749,1,  .749}}20.00                     &17.47    

                       &{\cellcolor[rgb]{.749,1,  .749}}30.30   &19.80        &{\cellcolor[rgb]{.749,1,  .749}}37.83       &27.77    \\ 
\hline
\multirow{3}{*}{Gynaeco.}           

& UCS                      &53.33                        &{\cellcolor[rgb]{.749,1,  .749}}61.53

      &{\cellcolor[rgb]{.749,1,  .749}}63.49          &60.60        &{\cellcolor[rgb]{.749,1,  .749}}81.81       &67.60 \\

& UCEC                       &{\cellcolor[rgb]{.749,1,  .749}}82.01                        &81.69                     &82.77              &{\cellcolor[rgb]{.749,1,  .749}}84.21                &{\cellcolor[rgb]{.749,1,  .749}}88.53       &86.99  \\

& CESC                      &64.03                        &{\cellcolor[rgb]{.749,1,  .749}}64.48  

                          &63.73        &{\cellcolor[rgb]{.749,1,  .749}}73.87          &82.24               &{\cellcolor[rgb]{.749,1,  .749}}82.35 \\

& OV                      &{\cellcolor[rgb]{.749,1,  .749}}64.93                          &52.17     
                   &{\cellcolor[rgb]{.749,1,  .749}}74.66        &70.42      &82.50                   &{\cellcolor[rgb]{.749,1,  .749}}83.95     \\ 
\hline

\multirow{2}{*}{Haematopoietic.}           
& THYM                      &91.76                          &{\cellcolor[rgb]{.749,1,  .749}}94.04                                               &{\cellcolor[rgb]{.749,1,  .749}}92.39         &91.56       &{\cellcolor[rgb]{.749,1,  .749}}94.54       &91.01  \\

& DLBC                     &22.22                          &{\cellcolor[rgb]{.749,1,  .749}}50.00  

                       &23.52              &{\cellcolor[rgb]{.749,1,  .749}}36.36     
                       &{\cellcolor[rgb]{.749,1,  .749}}60.86          &28.57      \\
\hline

\multirow{3}{*}{Liver, panc.}    

& CHOL           &{\cellcolor[rgb]{.749,1,  .749}}25.00                         &9.52

                &20.00              &{\cellcolor[rgb]{.749,1,  .749}}23.07       &34.78       &{\cellcolor[rgb]{.749,1,  .749}}38.46        \\

& LIHC                    &81.36                          &{\cellcolor[rgb]{.749,1,  .749}}84.24             
                     &85.09      &{\cellcolor[rgb]{.749,1,  .749}}85.34       &92.30       &{\cellcolor[rgb]{.749,1,  .749}}93.95                              \\

& PAAD      
& 58.18          &{\cellcolor[rgb]{.749,1,  .749}}66.12                            &68.42     &{\cellcolor[rgb]{.749,1,  .749}}74.79         &98.29       &90.90       \\ 

\hline

\multirow{2}{*}{Melanocytic malignancies} 
& SKCM                    &{\cellcolor[rgb]{.749,1,  .749}}96.02                       &94.31                                            &{\cellcolor[rgb]{.749,1,  .749}}97.57     &95.58      &{\cellcolor[rgb]{.749,1,  .749}}97.57       &94.91             \\

& UVM                    &78.87                          &{\cellcolor[rgb]{.749,1,  .749}}79.16                   
                   &{\cellcolor[rgb]{.749,1,  .749}}88.31                        &81.39     &{\cellcolor[rgb]{.749,1,  .749}}88.31       &80.85                 \\

\hline

\multirow{2}{*}{Prostate/testis}           & PRAD &91.37                    &{\cellcolor[rgb]{.749,1,  .749}}96.04                               &94.05          &{\cellcolor[rgb]{.749,1,  .749}}95.18         &98.29       &{\cellcolor[rgb]{.749,1,  .749}}98.85     \\

& TGCT &86.84                    &{\cellcolor[rgb]{.749,1,  .749}}93.69                               &90.58                     &{\cellcolor[rgb]{.749,1,  .749}}92.37        &97.32       &{\cellcolor[rgb]{.749,1,  .749}}98.24      \\ 
\hline

\multirow{3}{*}{Pulmonary}                 
& LUAD    &62.30                    &{\cellcolor[rgb]{.749,1,  .749}}67.89                             &{\cellcolor[rgb]{.749,1,  .749}}69.27                          &68.62           &{\cellcolor[rgb]{.749,1,  .749}}76.64       &75.71    \\

& LUSC   &61.65                    &{\cellcolor[rgb]{.749,1,  .749}}62.31                                 &61.08             &{\cellcolor[rgb]{.749,1,  .749}}64.51           &{\cellcolor[rgb]{.749,1,  .749}}72.72          &72.53     \\

& MESO &12.90                     &{\cellcolor[rgb]{.749,1,  .749}}50.90                                    &25.80        &{\cellcolor[rgb]{.749,1,  .749}}50.00         &52.63       &{\cellcolor[rgb]{.749,1,  .749}}65.11 \\ 
\hline

\multirow{4}{*}{Urinary tract}             
& BLCA     &75.88                    &{\cellcolor[rgb]{.749,1,  .749}}82.95                         
              &81.42     &{\cellcolor[rgb]{.749,1,  .749}}82.84          &{\cellcolor[rgb]{.749,1,  .749}}96.12       
              &94.84     \\
             
& KIRC &77.47                    &{\cellcolor[rgb]{.749,1,  .749}}82.46                                     &77.87                &{\cellcolor[rgb]{.749,1,  .749}}85.78         &{\cellcolor[rgb]{.749,1,  .749}}93.93       &91.83 \\

& KIRP &60.00                    &{\cellcolor[rgb]{.749,1,  .749}}62.94                                   &60.60                   &{\cellcolor[rgb]{.749,1,  .749}}63.67         &{\cellcolor[rgb]{.749,1,  .749}}88.64       &85.71   \\

& KICH &42.25                    &{\cellcolor[rgb]{.749,1,  .749}}52.27                                         &50.00        &{\cellcolor[rgb]{.749,1,  .749}}54.34           &{\cellcolor[rgb]{.749,1,  .749}}89.79       &87.23 \\ 
\bottomrule
\end{tabular}
\caption{F1-measure (in \%) for majority-3 search  through $k$-NN of the vertical search among 3,761 test WSIs for feature reduction effect by comparing top $500, 1000$, and $5000$ high variance features.}
\label{tab:Table2}
}

\end{table*}



\begin{table*}[t]
\small
\begin{tabular}{lccccccccccccc}
                & \multicolumn{3}{c}{SMA-T class }                                & \multicolumn{3}{c}{Negative Class }                             & \multicolumn{3}{c}{AMA Class }                                 & \multicolumn{3}{c}{SMA-V Class }    & \multicolumn{1}{c}{All}          \\ \hline
Method          & F1             & PR             & \multicolumn{1}{c|}{RE}             & F1             & PR             & \multicolumn{1}{c|}{RE}             & F1             & PR             & \multicolumn{1}{c|}{RE}             & F1             & PR             & \multicolumn{1}{c|}{RE}   & ACC             \\ \hline
Image-Level     & 16.67          & 100.00         & \multicolumn{1}{c|}{9.09}           & 88.00          & 81.15          & \multicolumn{1}{c|}{96.12}          & 89.89          & 90.90          & \multicolumn{1}{c|}{88.89}          & 66.67          & 73.68          & \multicolumn{1}{c|}{60.87}   & 81.95        \\
Patch-Level     & 00.00          & 00.00          & \multicolumn{1}{c|}{00.00}          & 79.67          & 68.53          & \multicolumn{1}{c|}{95.15}          & 84.71          & 90.00          & \multicolumn{1}{c|}{80.00}          & 23.53          & 36.36          & \multicolumn{1}{c|}{17.39}       & 69.27    \\
Multi-Scale     & 47.06          & 66.67          & \multicolumn{1}{c|}{36.36}          & 90.83          & 86.09          & \multicolumn{1}{c|}{96.12}          & 87.06          & 92.50          & \multicolumn{1}{c|}{82.22}          & 77.78          & 79.55          & \multicolumn{1}{c|}{76.09}     & 85.37      \\
RDMS \cite{dong2018reinforced}            & 55.56          & 71.43          & \multicolumn{1}{c|}{45.45}          & 93.00          & 95.88          & \multicolumn{1}{c|}{90.29}          & 91.49          & 87.76          & \multicolumn{1}{c|}{\textbf{95.56}} & 83.67          & 78.85          & \multicolumn{1}{c|}{89.13}  & 88.78           \\
SOS  \cite{maksoud2020sos}              & \textbf{70.00} & 71.43          & \multicolumn{1}{c|}{\textbf{63.64}} & \textbf{94.06} & \textbf{95.96} & \multicolumn{1}{c|}{92.23}          & 93.48          & 91.49          & \multicolumn{1}{c|}{\textbf{95.56}} & \textbf{85.42} & 82.00          & \multicolumn{1}{c|}{\textbf{89.13}}  & \textbf{90.73}     \\
D-SFV (Ours) & 66.67          & \textbf{85.71} & \multicolumn{1}{c|}{54.55}          & 93.84          & 91.67          & \multicolumn{1}{c|}{\textbf{96.12}} & \textbf{94.51} & \textbf{93.48} & \multicolumn{1}{c|}{\textbf{95.56}} & 84.44          & \textbf{86.36} & \multicolumn{1}{c|}{82.61}        & \textbf{90.73}       \\ \hline
\end{tabular}
\caption{Comparison of different WSI classification methods against Deep-SFV on Liver-Kidney-Stomach (LKS) dataset. F1, PR, and RE are in \%. }
\label{tab:Table3}
\end{table*}


The length of our proposed WSI  embedding is equal to the number of trainable parameters of the generative model, which is 1,407,105. Although here we employed a relatively small generative model for models with millions of parameters, the embeddings may not be suitable for efficient WSI search. The proposed gradients sparsity solves this issue by enforcing the generative model to use a smaller number of parameters for generating samples. In other words, by imposing sparsity on gradients, one can argue that the parameters that the gradients are zero w.r.t them do not have a significant contribution to generating those samples, so they can be removed from embeddings. We have validated the effect of the sparsity loss by selecting the gradients w.r.t. a subset of some parameters that leads to high-variance gradients per tissue type. \cref{tab:Table2} shows the feature reduction results for the same search with keeping $500, 5000$, and $40,000$ high variance bits. Based on \cref{tab:Table2} and \cref{fig:3} (left), keeping even $4000$ high variance bits not only outperforms Yottixel and C-GMM-FV in terms of search performance but also leads to significantly faster search speed. \cref{tab:Table2} clearly shows that the sparsity term helps the network to produce embedding with fewer but more informative bits. The right column in most subtypes outperforms the left column.  Based on the \emph{Fisher Vector Theory} this is intuitive that the gradients w.r.t generative model parameters show the contribution of each parameter to generating a sample. More precisely, by encouraging sparsity on the gradients, fewer parameters are contributing to generation; consequently, more parameters can be dropped in the final embedding.



\noindent \textbf{WSI Classification --} For this task, we validated the quality of obtained WSI embeddings on LKS datasets.  We trained a simple, fully connected network with two layers on top of the SFV embeddings for the purpose of WSI classification. The Liver-Kidney-Stomach (LKS) is the other publicly available dataset that we use for validating quality of WSI embeddings. The LKS dataset contains immunofluorescence WSIs realized by authors in~\cite{maksoud2020sos}. The dataset contains 684 WSIs from four classes Anti-Mitochondrial Antibodies (AMA),  Negative (Neg), Vessel-Type Anti-Smooth Muscle Antibodies (SMA-V), and
Tubule-Type Anti-Smooth Muscle Antibodies (SMA-T). This dataset contains one low-resolution image and also a set of patches per WSI. Following the same split in~\cite{maksoud2020sos}, we compared C-Deep-SFV against the proposed method in this paper Selective Objective Switch (SOS), Reinforced Dynamic Multi-Scale (RDMS),
\cite{dong2018reinforced} and three techniques for WSI classification, namely Image-Level, Patch-Level, and Conventional Multi-Scale (see~\cite{maksoud2020sos} for more detail). For this experiment, we only employed low-resolution images for training the backbone and then for each WSI we used is low-resolution image along with 5\% of high-resolution patches for training the CVAE and extracting the WSI embedding. The \cref{tab:Table3} presents the results in terms of precision, recall, F1, and accuracy measures where our method outperforms the Image-Level, Patch-Level, Conventional Multi-Scale, RDMS methods and achieves on par result compared with SOS. It is worth mentioning that our proposed architecture has one CNN backbone, while in SOS, two networks for low and high resolutions images have been used. Besides, embeddings obtained by or method are compact and suitable for WSI search.


\noindent \textbf{Ablation study --}
To study how gradient sparsity and quantization loss may affect retrieval performance, we conducted comprehensive ablation experiments.  \cref{fig:3} shows the average F1 measure across all sites for C-Deep-SFV and C-Deep-BFV where values \num{e-5},  \num{e-4}, and \num{e-3} have been tested for both $\lambda_4$ and $\lambda_5$. The average F1 decreases with increasing the $\lambda_4$. However, we should note that this experiment has been conducted with the same number of epochs (150). Our experiments showed that by increasing the $\lambda_4$ and also the number of epochs, the average F1 measure does not decrease. To see the effect of changing regularization parameters in more detail, we refer the readers to \cref{tab:Table4} in the supplementary material.


 \section{Conclusions}
\label{sec:conclusion}

We proposed a new framework based on deep conditional generative modelling and \emph{Fisher Vector Theory} for compact WSI representation. Unlike the common practice for WSI representation, i.e., MIL scheme, the training for the proposed method is instance-based, and as a result, GPU memory usage is the same as conventional training. Furthermore, we introduced new loss functions, gradient sparsity and gradient quantization for learning sparse and binary permutation-invariant representations, namely C-Deep-SFV and  C-Deep-BFV, suitable for efficient WSI retrievals. We showed that gradient sparsity loss function pushes the generative model to use parameters for generating a sample, and as a result, one can reduce the dimensionality of the WSI embeddings and still achieve a good performance.  
The WSIs representations were validated on the largest public archive of WSIs, The TCGA WSIs and also the LKS dataset for both WSI search and classification tasks. The proposed method outperforms \emph{Yottixel} a recent search engine for histopathology images and \emph{GMM-based Fisher Vector}. Furthermore, we also achieved competitive results against state-of-the-art in WSI classifications on both lung and LKS public benchmark datasets.

 \section{Data availability}
\label{sec:Data_availability}

The diagnostic slides from The Cancer Genomic Atlas (TCGA) repository ~\cite{weinstein2013cancer} dataset is available \href{https://portal.gdc.cancer.gov/}{here}. The Liver Kidney Stomach (LKS) ~\cite{maksoud2020sos} is available here in \href{https://github.com/cradleai/LKS-Dataset}{this repository}.

{
    \small
    \bibliographystyle{ieee_fullname}
    \bibliography{macros,main}
}

\appendix
\setcounter{page}{1}

\twocolumn[
\centering
\Large
\vspace{0.5em}Supplementary Material \\
\vspace{1.0em}
] 
\appendix

\section{Detailed ablation study}
\begin{table*}
\small
\centering
\scriptsize
\begin{tabular}{cccc|cc|cc}
\toprule
  &     &   & \multicolumn{3}{c}{F1-measure (in \%)}    \\\cline{3-8}
  
   &   & $\lambda_4=$\num{e-5}  & $\lambda_5=$\num{e-5}  & $\lambda_4=$\num{e-4}   & $\lambda_5=$\num{e-4} & $\lambda_4=$\num{e-3}  & $\lambda_5=$\num{e-3}      \\
       
Site & Subtype  & C-Deep-SFV  & C-Deep-BFV  & C-Deep-SFV  & C-Deep-BFV & C-Deep-SFV  & C-Deep-BFV    \\ 
\toprule

\multirow{2}{*}{Brain}                 

& LGG                     &90.88          &92.87                   &93.07            &93.10                          &92.72          &93.80                    \\

&GBM          &91.67                         &93.91                &93.76
                &93.99                         &93.53         &94.70  \\
\hline

\multirow{3}{*}{Endocrine}                 
& THCA                   &98.48                         &97.29                    &99.24               &99.24                        &98.74       &98.75             \\

&ACC                       &92.55                         &91.30                          &95.08               &94.62                           &95.13      &94.62            \\
                 
& PCPG                   &88.40                         &90.07            &92.95          &90.64                         &91.42             &90.37          \\ 
\hline

\multirow{4}{*}{Gastro.}    
& ESCA                     &69.42
                    &67.76                      &65.42                    &75.43                         &77.04     &70.17  \\

& COAD                     &73.38                          &75.97                       
&74.93                         &74.44             &76.21    &74.01    \\

& STAD                     &84.07                          &79.23   
&80.75                       &83.48                          &84.45       &84.01      \\

& READ                     &25.49                     &33.33    

&31.77                &30.30                        &30.18              &29.90          \\ 
\hline
\multirow{3}{*}{Gynaeco.}           

& UCS                      &81.15                        &74.28

&72.13              &74.62                       &73.23         &71.64\\

& UCEC                       &89.97                        &86.18            &87.29              &83.33                   &89.42            &85.58        \\

& CESC                      &82.72                        &84.30  

&81.44                &70.47                    &86.84    &74.17\\

& OV     &79.48                          &73.68     
&83.95                   &77.33                        &79.01       &81.01         \\ 
\hline

\multirow{2}{*}{Haematopoietic.}           
& THYM                      &94.54                          &93.49                            &93.97         &96.93                         &93.56         &95.23  \\

& DLBC                     &60.86                          &42.10 

&54.54                    &80.00                  &35.29              &60.00\\
\hline

\multirow{3}{*}{Liver, panc.}    

& CHOL           &41.66                         & 25.00

&48.27                &41.66                       &14.81           &16.66       \\

& LIHC                    &94.91                         &92.66            
&93.91                      &94.00                       &88.66      &92.35                                   \\

& PAAD      
&91.97         &90.90         &94.65                &93.93                           &86.82    &93.12              \\ 

\hline

\multirow{2}{*}{Melanocytic malignancies} 
& SKCM                    &98.13                       &98.37                       &98.37                &98.92                        &97.23     &97.86 \\

& UVM                    &90.41                          &92.30                   
&92.30                        &94.73                      &88.37                        &89.18            \\

\hline

\multirow{2}{*}{Prostate/testis}           
&PRAD     &99.42                   &99.14                      &98.55        
&99.14            &98.56            &98.85            \\

& TGCT &99.11                    &98.66                     &97.81                   &98.67          &98.56                   &98.23  \\ 
\hline

\multirow{3}{*}{Pulmonary}                 
& LUAD &77.84                    &75.84                   &79.12               &74.88          &76.95                       &78.70             \\

& LUSC      &72.67                    &74.20                        &72.95                 &72.04            &71.72          &72.20   \\

& MESO          &63.63                     &50.00                           &51.28                   &39.99                  &72.72        &50.00\\ 
\hline

\multirow{4}{*}{Urinary tract}             
& BLCA     &94.20                    &94.62                         
&92.83                        &94.20               &96.14     &94.02            \\
             
& KIRC &92.73                    &90.90                            &93.75            &91.47         &93.50              &89.35  \\

& KIRP      &86.69                    &83.94                         &84.01          &84.05            &87.63               &78.41   \\

& KICH      &90.52                    &90.32                        &89.58           &89.36                &90.72         &
87.64\\ 
\bottomrule
\end{tabular}
\caption{Ablation study on $\lambda_4$ and $\lambda_5$ regularization parameters based on Majority-3 search  through $k$-NN of the vertical search among 3761 test WSIs.}
\label{tab:Table4}
\end{table*}

The Table \ref{tab:Table4} shows the detailed ablation study on the gradient sparsity and quantization losses regularization parameters. More precisely, The F1 measure across all sites for C-Deep-SFV and C-Deep-BFV with regularization parameters equal to  \num{e-5},  \num{e-4}, and \num{e-3}for  $\lambda_4$ and $\lambda_5$ are reported in Table \ref{tab:Table4}.



\section{Full names for cancer subtypes}

The full description of the abbreviations for cancer subtypes used in this paper have been presented in Table \ref{tab:Table5}.
\begin{table}[t]
    \centering
   \scriptsize
    \begin{tabular}{llc}
    Abbreviation &                                  Primary Diagnosis  \\
    \hline
    ACC                      &                           Adrenocortical Carcinoma  \\
    BLCA                     &                       Bladder Urothelial Carcinoma  \\
    CESC                     &  Cervical Squamous Cell Carcinoma and Endocervical Adenoc.   \\
    CHOL                     &                                 Cholangiocarcinoma  \\
    COAD                     &                               Colon Adenocarcinoma  \\
    ESCA                     &                               Esophageal Carcinoma  \\
    GBM                      &                            Glioblastoma Multiforme  \\
    KICH                     &                                 Kidney Chromophobe \\
    KIRC                     &                  Kidney Renal Clear Cell Carcinoma  \\
    KIRP                     &              Kidney Renal Papillary Cell Carcinoma  \\
    LGG                      &                           Brain Lower Grade Glioma  \\
    LIHC                     &                     Liver Hepatocellular Carcinoma  \\
    LUAD                     &                                Lung Adenocarcinoma \\
    LUSC                     &                       Lung Squamous Cell Carcinoma  \\
    MESO                     &                                       Mesothelioma \\
    OV                       &                  Ovarian Serous Cystadenocarcinoma  \\
    PAAD                     &                          Pancreatic Adenocarcinoma  \\
    PCPG                     &                 Pheochromocytoma and Paraganglioma  \\
    PRAD                     &                            Prostate Adenocarcinoma  \\
    READ                     &                              Rectum Adenocarcinoma  \\
    STAD                     &                             Stomach Adenocarcinoma  \\
    TGCT                     &                        Testicular Germ Cell Tumors  \\
    THCA                     &                                  Thyroid Carcinoma  \\
    UCS                      &                             Uterine Carcinosarcoma  \\
    \bottomrule
\end{tabular}
\caption{Full description for primary diagnosis abbreviations used in the paper.} 
\label{tab:Table5}
    \end{table}


\end{document}